%% file: acl_latex.tex
\pdfoutput=1
\documentclass[11pt]{article}

\usepackage{hyperref}
\usepackage[]{acl}
\usepackage{dsfont}
\usepackage{times}
\usepackage{latexsym}
\usepackage{algorithm}
\usepackage{algcompatible,amsmath}
\usepackage{graphicx}
\usepackage{url}
\usepackage{multirow}
\usepackage{comment}
\usepackage{amssymb}
\usepackage{pifont}
\usepackage{balance}
\usepackage{mathtools}
\usepackage{xcolor}
\usepackage{svg}
\usepackage{colortbl}
\usepackage{bbm}
\usepackage{enumitem}
\usepackage{marvosym}
\usepackage{textcomp}
\usepackage{array}
\usepackage{booktabs}
\usepackage{tabularx}
\usepackage{CJKutf8}
\usepackage{algorithm}  
\usepackage{algorithmicx}  
\usepackage{algpseudocode} 
\usepackage{subfig}
\usepackage{tcolorbox}
\usepackage{cleveref}
\usepackage[T1]{fontenc}

\usepackage[utf8]{inputenc}
\usepackage{fdsymbol}

\usepackage{microtype}

\usepackage{inconsolata}
\providecommand{\tightlist}{%
  \setlength{\itemsep}{0pt}\setlength{\parskip}{0pt}}

\newlist{researchquestions}{enumerate}{1}
\setlist[researchquestions]{label*=\textbf{RQ\arabic*}}
\title{\textsc{ProxyQA}: An Alternative Framework for Evaluating Long-Form\\ Text Generation with Large Language Models}

\author{Haochen Tan$^{\clubsuit\varheartsuit\dagger}$\ Zhijiang Guo$^{\spadesuit\dagger}$, Zhan Shi$^{\diamondsuit}$, Lu Xu$^{\spadesuit}$, Zhili Liu$^{\spadesuit,\heartsuit}$ Yunlong Feng$^{\triangleleft}$ \\ {\bf Xiaoguang Li}$^{\spadesuit}$, {\bf Yasheng Wang}$^{\spadesuit}$, {\bf Lifeng Shang}$^{\spadesuit}$, {\bf Qun Liu}$^{\spadesuit}$, {\bf Linqi Song}$^{\clubsuit \varheartsuit}\thanks{~~Corresponding author.}$ \\
$^{\clubsuit}$City University of Hong Kong
$^{\spadesuit}$Huawei Noah's Ark Lab 
$^{\diamondsuit}$Huawei Hisilicon \\
$^{\heartsuit}$Hong Kong University of Science and Technology 
$^{\triangleleft}$Harbin Institute of Technology\\
$^{\varheartsuit}$City University of Hong Kong Shenzhen Research Institute\\
\texttt{haochetan2-c@my.cityu.edu.hk}\hspace{1em}
\texttt{guozhijiang@huawei.com}\\
\texttt{linqi.song@cityu.edu.hk}
}

\begin{document}
\maketitle
\def\thefootnote{$\dagger$}\footnotetext{These authors contributed equally}

\input{sections/abstract}
\input{sections/intro}
\input{sections/related}
\input{sections/dataset}
\input{sections/experiment}
\input{sections/results}
\input{sections/analysis}
\input{sections/conclusion}

\input{sections/limitation}

\bibliography{anthology,custom}

\appendix

\input{sections/appendix}

\end{document}

%% file: sections/abstract.tex
\begin{abstract}
Large Language Models (LLMs) have succeeded remarkably in \textbf{understanding} long-form contents. However, exploring their capability for \textbf{generating} long-form contents, such as reports and articles, has been relatively unexplored and inadequately assessed by existing benchmarks. The prevalent evaluation methods, which predominantly rely on crowdsourcing, are recognized for their labor-intensive nature and lack of efficiency, whereas automated metrics, such as the ROUGE score, demonstrate discordance with human judgment criteria. In this paper, we propose \textsc{ProxyQA}, an innovative framework dedicated to assessing long-text generation. \textsc{ProxyQA} comprises in-depth human-curated \textit{meta-questions} spanning various domains, each accompanied by specific \textit{proxy-questions} with pre-annotated answers. LLMs are tasked to generate extensive content in response to these meta-questions, by engaging an evaluator and incorporating the generated texts as contextual background, \textsc{ProxyQA} assesses the generated content's quality through the evaluator's accuracy in addressing the \textit{proxy-questions}.
We examine multiple LLMs, emphasizing \textsc{ProxyQA}'s demanding nature as a high-quality assessment tool. Human evaluation demonstrates that the \textit{proxy-question} method is notably self-consistent and aligns closely with human evaluative standards.
The dataset and leaderboard is available at \url{https://proxy-qa.com}.
\end{abstract}

%% file: sections/intro.tex
\section{Introduction}
Recent Large Language Models (LLMs) have made significant advancements~\citep{BrownMRSKDNSSAA20,touvron2023,llama2,chatgpt,gpt4}. GPU technology innovations and memory-efficient attention mechanisms~\citep{DaoFERR22, Dao2023} have further enabled LLMs to model context sequences spanning tens of thousands of tokens~\citep{claude200k, gpt4128k}, paving the way for sophisticated applications such as analyzing complex scientific essays and generating detailed reports. As long-context LLMs evolve, several benchmarks have emerged to evaluate their ability to handle extensive contexts \citep{shaham2023, Bai2023, An2023, zhang2023infinitebench}. However, these assessments primarily focus on LLMs' \textbf{comprehension} of lengthy passages, using automated metrics to measure performance. This leaves a significant gap in understanding LLMs' proficiency in \textbf{generating} long-form texts, an essential aspect that requires further investigation.

\begin{figure}
   \centering
   \includegraphics[scale = 0.6]{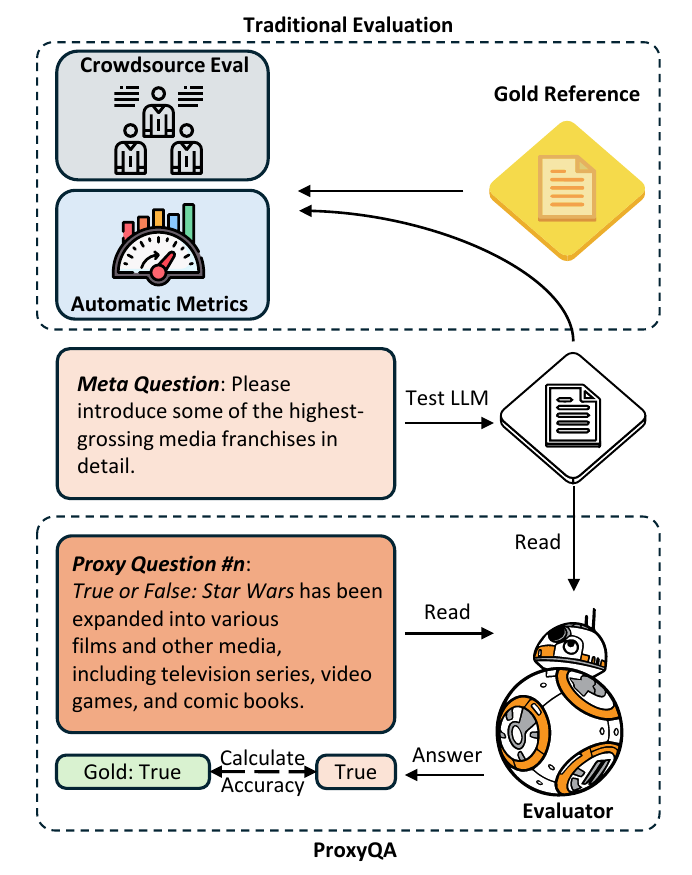}
   \caption{Prior efforts assess generated content by matching it with references through human evaluation or automated metrics. \textsc{ProxyQA} evaluates the knowledge coverage and informativeness by checking if generated contents contain sufficient information to answer a set of proxy questions.}
   \label{fig:framework}
\end{figure}

One primary roadblock to understanding LLMs' capability to generate long-form texts is the lack of competent evaluation methods. Current methods, often involving a combination of automated metrics and crowdsourced annotations, leave much to be desired~\citep{XuSIC23}. For instance, automated metrics that use word-level string~\citep{lin2004} or meaning representation matching~\citep{YuanNL21} rely on gold references, which unfortunately do not exist for many generation tasks, such as reports or essay writing. Furthermore, these automated metrics are inadequate for reliably assessing long-form content due to the considerable and unstructured space of potential outputs~\citep{Celikyilmaz2020,krishna2021}.  Human evaluation has its own set of issues too. Crowdsourced workers may lack the necessary expertise for evaluating knowledge-rich content, and domain experts' subjective preferences could result in inconsistent evaluations~\citep{XuSIC23}. While recent studies explored using LLMs for evaluation~\citep{ChiangL23,LiuIXWXZ23}, LLMs have been found to lack the most current information required for precise verification. Moreover, their assessments have been observed to be inconsistent~\citep{ShenCNYB23}. There is a clear need for more robust and precise evaluation methods.

To address this issue, we introduce \textsc{ProxyQA}, a benchmark comprising human-curated meta-questions covering a wide range of subjects, from computer science to history. These meta-questions require domain expertise and up-to-date knowledge, prompting LLMs to generate detailed and comprehensive responses. To assess knowledge coverage and informativeness, we pair each meta-question with a series of proxy-questions and answers that capture its essential points. As illustrated in Figure~\ref{fig:framework}, \textsc{ProxyQA} uses an evaluator to answer proxy-questions based on the long-form content produced by LLMs, rather than comparing the output to a reference. If the generated content is sufficiently detailed and accurate, it should equip the evaluator with enough information to thoroughly answer all associated proxy-questions.

\textsc{ProxyQA} offers several benefits.  By employing proxy questions and an evaluator, it eliminates the need for direct comparison against a single gold reference, enabling a more accessible and subjective evaluation. Using this approach allows evaluators without specific domain knowledge to assess content. Additionally, unlike previous datasets compiled from online sources~\citep{NguyenRSGTMD16,FanJPGWA19} that potentially leading to data contamination~\citep{SainzCGELA23}, all the \textit{proxy-questions} and answers are invisible to public, thereby preventing data leakage. We apply \textsc{ProxyQA} to extensively test different LLMs~\citep{touvron2023,llama2,alpaca,ChiangL23, chatgpt,gpt4}, including the LLMs that enhanced iterative reasoning~\citep{YaoZYDSN023} and retrieval augmentation~\citep{new_bing,geminiteam2023gemini}. 
A systematic human evaluation demonstrates that \textsc{ProxyQA} offers a highly consistent evaluation scope, surpassing inter-human agreement rates while maintaining strong correlations with the majority preferences of humans. 
In-depth analysis shows that ProxyQA effectively eliminates the bias that LLM-as-a-judge~\citep{zheng2023judging} prefer content generated by GPT models, and the evaluators' accuracy on the proxy-questions accurately reflects the quality of the generated contents.

%% file: sections/related.tex
\begin{figure*}
    \centering
    \includegraphics[width=.9\linewidth]{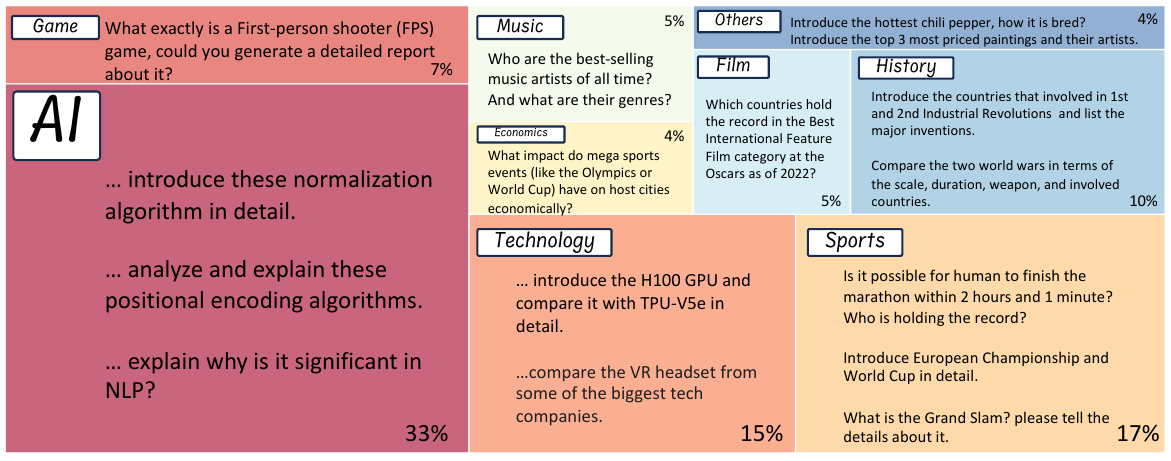}
    \caption{Meta-questions in \textsc{ProxyQA} cover various domains, such as AI research, historical event investigations, sports and entertainment analysis, and more.}
    \label{fig:example}
\end{figure*}

\section{Related Work}

\subsection{Long-Form Text Generation}

Significant strides have been made in long-form text generation, particularly in story generation~\citep{FanLD19, XuPSPFAC20}, paragraph completion~\citep{KangH20}, long-term conversation~\citep{XuSW22} and article generation~\citep{HuaW20,HuCLXWH22}. A closely related field is long-form question answering~\citep{FanJPGWA19,DasigiLBCSG21,StelmakhLDC22,Lee2023}, which involves generating detailed responses to complex information-seeking questions. ELI5~\citep{FanJPGWA19} was a pioneer dataset for generating explanatory paragraphs in response to open-ended questions, utilizing answers from Reddit.
QASPER~\citep{DasigiLBCSG21} and QASA~\citep{Lee2023} extend general factoid questions to the domain of scientific literature.  
Evaluating answers on these datasets relies on comparing the generated texts with the provided single reference. However, open-ended questions can be answered in myriad different ways. 
ASQA~\citep{StelmakhLDC22} also introduces a set of disambiguated questions from AmbigQA~\citep{MinKLLYKIZH23} for evaluating ambiguous questions. They assume that long-form answers to ambiguous questions should resolve ambiguity. In contrast, ProxyQA aims to gauge the informativeness and comprehensiveness of long-form answers, without being confined solely to ambiguous questions.

\subsection{Text Generation Evaluation}

Automated metrics such as surface form matching~\citep{lin2004,BanerjeeL05} and semantic representation comparison~\citep{ZhangKWWA20,YuanNL21}, face challenges with long-form content due to their inability to handle the diversity of the potential outputs~\citep{Celikyilmaz2020,krishna2021}. They often do not align with human judgment~\citep{XuSIC23}. Attempts to use LLMs for evaluation~\citep{LiuIXWXZ23,ChiangL23,Liu2024aligning} are hindered by their limited access to current information and inconsistency in performance~\citep{ShenCNYB23}. Evaluators also have difficulties, particularly if they lack expertise, which can impair their judgment on key dimensions like informativeness and factuality~\citep{gillick2010,iskender2020}. Strategies to enhance human evaluations include A/B testing, as seen with HURDLES~\citep{krishna2021} and WebGPT~\citep{Nakano2021WebGPTBQ}, with the latter demonstrating that providing evidence helps annotators make more informed decisions. The list of proxy questions in \textsc{ProxyQA} can be viewed as evidence to assist the evaluator in making decisions. While some research focused on coherence~\citep{goyal2022,jiang2022, DengKR22} and factuality~\citep{goyal2020,laban2022, MinKLLYKIZH23} in related tasks like summarization, our work emphasizes informativeness and coverage.

%% file: sections/dataset.tex
\section{A Long-form Generation Benchmark}
An alternative framework for evaluating long-form text generation is created. 100 meta-questions are annotated to prompt LLMs to generate detailed and informative responses, which cover subjects in artificial intelligence (AI) research, historical event investigations, sports and entertainment event analysis. The topic distribution is shown in Figure~\ref{fig:example}. Each meta-question accompanies various \textit{proxy-questions} with annotated answers, which are invisible to the LLMs to be tested. To proxy the objective evaluation of the generated contents to subjective metrics, we adopt an \textit{evaluator} that takes the generated contents as contextual background and answers the proxy-questions. We assume that only if the contextual background is informative and comprehensive, can the proxy-questions be well-addressed by the evaluator. Therefore, the quality of the generated contents is reflected in the evaluator's accuracy on the proxy-questions. 

\subsection{Creating the Meta-questions}
\label{para:domain}
Meta-questions were manually raised by five experienced researchers, who were instructed to initiate meta-questions in areas with which they were most familiar or had a keen interest. Besides the most-concerned topics such as AI research, sports and gaming, less popular domains such as infrastructure and agriculture are also included. We focus on questions that are aligned with real-life scenarios and should be well-addressable in reports or articles. For instance, a pertinent question within the \textit{Computer Science} domain could be: ``Could you elaborate on the development of Model Parallelism and Pipeline Parallelism, detailing key milestones and contributions?'' This meta-question aligns with interests in parallel computing techniques. In contrast, questions of the sort ``Did Aristotle use a laptop?'' from StrategyQA~\citep{geva-etal-2021-aristotle} are omitted due to their lack of occurrence in realistic settings.

\paragraph{Difficulty of Meta-question}
Meta-questions can be classified into two levels of difficulty: easy and hard. The annotation process was bifurcated; half of the experts annotated questions using only Wikipedia, denoting easy questions, while the rest utilized various resources to labeling difficult questions. This method was adhered to in subsequent annotation rounds, with experts verifying knowledge sources to accurately categorize question difficulty. Easy questions can be sufficiently answered using only information from Wikipedia, while hard questions demand the integration of Wikipedia content with insights derived from a wider range of open-domain knowledge sources to formulate a comprehensive response. Generally, most LLMs can effectively address easy questions given their extensive training on Wikipedia corpora. Conversely, hard questions pose a challenge to the models' ability to acquire information beyond the commonly used pre-training data, necessitating access to specialized private corpora, web searches, or document retrieval for a comprehensive response. For more details, please refer to Appendix~\ref{appendix:dist}. 

\subsection{Annotation of Proxy-Question}
The evaluation of the generated reports is proxied to the evaluator's accuracy on proxy-questions. Experts are tasked with identifying the pivotal content that a satisfactory answer to a meta-question must contain. Then they craft a series of proxy-questions that probe these identified key points. For instance, regarding the example of model parallelism and pipeline parallelism mentioned in Section~\ref{para:domain}, a thorough answer should incorporate in-depth information about Gpipe~\cite{gpipe}, Megatron~\cite{megatron}, and other pertinent subjects.
Therefore, annotators develop proxy-questions that specifically focus on Gpipe, Megatron and other related topics. As our destination is not to stress-test the evaluator but to quantify the quality of the generated contents, we present straightforward and concise proxy-questions, deliberately avoiding multi-hop and complex reasoning queries. Each annotated response is provided in a boolean format, ensuring that evaluators can effortlessly answer these proxy-questions, given a sufficiently high-quality generated context. Conversely, if an evaluator struggles to address these simple proxy-questions based on the provided context, it indicates a significant deficiency in the generated context, with crucial information being absent.

\subsection{Iterative Annotation}
As the foundation of the evaluation in \textsc{ProxyQA}, proxy-questions are curated through a multi-round annotation process, where experts iteratively exchange the meta-questions while supplementing and verifying proxy-questions annotated by the others. 
Each meta-question is thus repeatedly given to different experts to label different proxy-questions, until a consensus is reached that all experts agree that the points covered by the proxy-questions are sufficiently comprehensive. 
Such an alternate labeling process ensures a multi-perspective rubric, leading to an experts-consolidated benchmark. After the iterative annotation, each meta-question is coupled with 15.5 proxy-questions on average, we then measure the inter-annotator agreement of \textsc{ProxyQA}. We randomly extracted a subset of 50 proxy-questions and tasked the annotators to re-annotate them. Following the ~\citet{kazemi-etal-2021-claim}, the Randolph's free-marginal multi-rater $\kappa$~\cite{kappa}, an alternative to Fleiss' $\kappa$ are measured. \textsc{ProxyQA} achieves $\kappa = 0.936$ thanks for the introduction of the iterative annotation process. 

\subsection{Quality Assurance}
\label{sec:quality}
The meta-question and its corresponding proxy-questions were annotated and quality-checked iteratively. 
During the annotation, we excluded three meta-questions that were offensive, politically sensitive, ethically concerning, or not safe for work (NSFW). Meta-questions that have been previously posted and well-addressed on relevant forums are replaced. This ensures that LLMs cannot generate answers by directly copying content from these platforms, therefore preserving the integrity of our dataset. 
Additionally, it is worth noting that real-world knowledge evolves over time. Therefore, to ensure the quality of \textsc{PrxoyQA}, our experts are required to review and update proxy-questions periodically, as detailed in the Appendix~\ref{appendix:mt}.
\input{tables/main_results}

\subsection{Evaluators}
To ensure the generated contents such as markdown tables and math formulas can be well encoded, GPT-4 and its variant, GPT-4-Turbo (GPT-4-1106-Preview), are utilized as evaluators. Instead of applying retrieval augmented generation, evaluators are required to read the generated contents and answer the proxy-questions. We prompt the evaluator to formulate answers to the proxy-questions strictly from the information presented within the contextual background. Evaluator's accuracy can be high only if the generated contextual background is informative and comprehensive enough. The objective assessment of these reports is subsequently anchored to the precision of the GPT evaluator. 
Moreover, we take the mean value of evaluation results from GPT-4 and GPT-4-Turbo to reinforce the reliability and robustness of the assessment.

%% file: tables/main_results.tex
\begin{table*}[t]
\centering

\scalebox{0.84}{

\begin{tabular}{lllllllllllll}
\toprule[0.8pt]
\multicolumn{1}{c}{\textbf{}} & \multicolumn{1}{c}{} & \multicolumn{3}{c}{\textbf{GPT-4}}                                                                                                   &                                        & \multicolumn{3}{c}{\textbf{GPT-4-Turbo}}                                                                                             &                                        & \multicolumn{3}{c}{\textbf{Average}}                                                                                                \\ \cline{3-5} \cline{7-9} \cline{11-13} 
\multicolumn{1}{c}{}          & \multicolumn{1}{c}{} & \multicolumn{1}{c}{\textit{\textbf{Easy}}} & \multicolumn{1}{c}{\textit{\textbf{Hard}}} & \multicolumn{1}{c}{\textit{\textbf{avg.}}} & \multicolumn{1}{c}{\textit{\textbf{}}} & \multicolumn{1}{c}{\textit{\textbf{Easy}}} & \multicolumn{1}{c}{\textit{\textbf{Hard}}} & \multicolumn{1}{c}{\textit{\textbf{avg.}}} & \multicolumn{1}{c}{\textit{\textbf{}}} & \multicolumn{1}{c}{\textit{\textbf{Easy}}} & \multicolumn{1}{c}{\textit{\textbf{Hard}}} & \multicolumn{1}{c}{\textit{\textbf{avg.}}} \\ \hline
\multicolumn{13}{c}{\textit{Base LLaMA}}                                                                                                                                                                                                                                                                                                                                                                                                                                                                                                                 \\ \hline
LLaMA-7B                      &                      & 5.25                                       & 0.68                                       & 3.05                                       &                                        & 5.89                                       & 1.23                                       & 3.64                                       &                                        & 5.57                                       & 0.96                                       & 3.34                                       \\
LLaMA2-7B                    &                      & 4.74                                       & 0.55                                       & 2.72                                       &                                        & 5.38                                       & 0.55                                       & 3.04                                       &                                        & 5.06                                       & 0.55                                       & 2.88                                       \\
LLaMA2-13B                   &                      & \textbf{6.15}                              & \textbf{1.65}                              & \textbf{3.97}                              &                                        & \textbf{6.91}                              & \textbf{1.37}                              & \textbf{4.24}                              &                                        & \textbf{6.53}                              & \textbf{1.51}                              & \textbf{4.11}                              \\ \hline
\multicolumn{13}{c}{\textit{Instruction-Finetuned LLaMA}}                                                                                                                                                                                                                                                                                                                                                                                                                                                                                                   \\ \hline
Alpaca-7B                     &                      & 12.42                                      & 5.62                                       & 9.14                                       &                                        & 14.60                                      & 9.33                                       & 12.05                                      &                                        & 13.51                                      & 7.48                                       & 10.60                                      \\
Vicuna-13B                    &                      & 19.85                                      & \textbf{17.15}                             & 18.54                                      &                                        & \textbf{22.66}                             & \textbf{21.26}                             & \textbf{21.99}                             &                                        & 21.25                                      & \textbf{19.20}                             & \textbf{20.26}                             \\
LLaMA2-7B-Chat               &                      & \textbf{21.25}                             & 16.74                                      & \textbf{19.07}                             &                                        & 20.23                                      & 18.11                                      & 19.20                                      &                                        & 20.74                                      & 17.42                                      & 19.14                                      \\
LLaMA2-13B-Chat              &                      & 21.13                                      & 16.87                                      & \textbf{19.07}                             &                                        & 22.02                                      & 17.42                                      & 19.80                                      &                                        & \textbf{21.57}                             & 17.15                                      & 19.44                                      \\ \hline
\multicolumn{13}{c}{\textit{GPT APIs}}                                                                                                                                                                                                                                                                                                                                                                                                                                                                                                                    \\ \hline
GPT-3.5-Turbo                 &                      & 25.61                                      & 21.40                                      & 23.57                                      &                                        & 26.12                                      & 22.36                                      & 24.30                                      &                                        & 25.87                                      & 22.97                                      & 23.94                                      \\
GPT-4                         &                      & 30.35                                      & 23.05                                      & 26.82                                      &                                        & 30.35                                      & 23.55                                      & 27.55                                      &                                        & 30.35                                      & 23.80                                      & 27.19                                      \\
GPT-4-Turbo                   &                      & \textbf{35.21}                             & \textbf{31.69}                             & \textbf{33.50}                             &                                        & \textbf{34.83}                             & \textbf{33.88}                             & \textbf{34.37}                             &                                        & \textbf{35.02}                             & \textbf{32.78}                             & \textbf{33.94}                             \\ \hline
\multicolumn{13}{c}{\textit{Web-Augmented LLMs}}                                                                                                                                                                                                                                                                                                                                                                                                                                                                                                            \\ \hline
ReAct (GPT-4)               &                      & 20.74                                      & 13.72                                      & 17.35                                      &                                        & 20.49                                      & 13.17                                      & 16.95                                      &                                        & 20.61                                      & 13.44                                      & 17.15                                      \\
ReAct (GPT-4-Turbo)         &                      & 23.94                                      & 18.11                                      & 21.13                                      &                                        & 23.56                                      & 18.79                                      & 21.26                                      &                                        & 23.75                                      & 18.45                                      & 21.19                                      \\
Bard (Gemini Pro)            &                      & 26.63                             & 22.22                             & 24.50                            &                                        & 25.48                             & 25.51                             & 25.50                             &                                        & 25.06                            & 23.87                             & 25.00                             \\
New Bing (Creative Mode)        &                      & \bf39.56                                      & \bf37.72                                      & \bf38.67                                      &                                        & \bf40.33                                      & \bf39.78                                      & \bf40.06                                      &                                        & \bf39.95                                      & \bf38.75                                      & \bf39.37                                      \\
\bottomrule[0.8pt]
\end{tabular}
}
\caption{Evaluation of various LLMs' knowledge coverage and informativeness on \textsc{ProxyQA}, accuracy on the easy and hard splits are reported.}
\label{tab:main_result}
\end{table*}

%% file: sections/experiment.tex
\section{Experiments}
\subsection{Setup}
\paragraph{Assessment and Models}
To measure LLMs' performance on \textsc{ProxyQA}, we compute the accuracy of the proxy-questions across easy and hard splits. Open-sourced LLMs such as LLaMA and its instruction-finetuned variants are tested. We also evaluated closed-sourced LLMs (e.g. GPT) and web-augmented LLMs, including Bard (Gemini Pro)~\cite{geminiteam2023gemini} and New Bing~\cite{new_bing}. Details of all the tested models are in the Appendix~\ref{appendix:baseline}. Each LLM is prompted with ``Write a well-structured and extensive report to answer the question: [META-QUESTION]'' under the setting of zero-shot evaluation. The max decoding length of open-sourced and closed-sourced LLMs is set to their reported maximum length. All the other hyperparameters of decoding strategies are the same as their reported settings in the original paper or API documentation.

\paragraph{Setting of the Evaluator}
We feed GPT-4 and GPT-4-Turbo with the prompt shown in Appendix~\ref{appendix:prompt}. Both evaluators used the same decoding strategy with top\_p$=$1, max\_tokens$=$10, and frequency\_penalty$=$0. Each evaluator only processes one sub-questions at a time.
We calculate the evaluator's average accuracy to represent the generated content's overall quality.

\paragraph{Research Questions}
To carry out an in-depth investigation of long-form content generation and the effectiveness of our proposed \textsc{ProxyQA}, the following research questions are naturally raised:\\
   \textbf{RQ1.} How do open-sourced LLMs compare to proprietary models' ability to generate extensive reports or articles? (\Cref{sec:main})\\
   \textbf{RQ2.} How well do the modern LLMs grasp the knowledge with different difficulty levels? (\Cref{sec:diff_r}) \\
   \textbf{RQ3.} How do LLMs perform in generating long content from different domains? (\Cref{sec:domain_r})\\
   \textbf{RQ4.} How does the iterative and alternate annotation of proxy-questions affect the performance of the LLMs? (\Cref{sec:alter_anno})\\
   \textbf{RQ5.} Does \textsc{ProxyQA} give higher ratings for generated content with longer length? (\Cref{sec:gen_len})

%% file: sections/results.tex
\subsection{Main Results}
\label{sec:main}
\textit{The ability of open-sourced LLMs to generate comprehensive and extensive content is far behind the proprietary models.}
As shown in Table~\ref{tab:main_result}, the base versions of the LLaMAs series demonstrate a limited capacity to produce long-form content. However, notable enhancements are achieved through instruction-based supervised fine-tuning (SFT), as the Vicuna-13B and LLaMA2-13B-Chat transcend most other open-sourced models, evidencing their superior capability in delivering acceptable content. However, compared to the proprietary models, the open-source LLMs far lag behind GPT models. Even the GPT-3.5-Turbo outperforms the entire suite of open-source LLMs by a significant margin, and GPT-4-Turbo maintains a substantial lead. This underscores the considerable gap that open-source LLMs must bridge to match the performance of their proprietary counterparts.



\subsection{Results on Different Levels of Difficulty}
\label{sec:diff_r}
\textit{Well-pretrained LLMs surpass others on all fronts but struggle with hard questions, while well-designed retrieval-augmented generation (RAG) significantly make up the shortcoming.}
Table~\ref{tab:main_result} illustrates a pronounced decline in performance among most large language models (LLMs) on the more challenging subset of questions, which cannot be well-solved solely with information from Wikipedia. Notably, even the powerful GPT-4 exhibits a marked decrease in efficacy($6.55\downarrow$).  
In contrast, equipped with the GPT-4 with a search engine and well-designed searching strategy, New Bing Creative Mode performs more robustly than other LLMs, exhibiting a comparatively minor performance loss( $39.95\rightarrow $38.75). However, RAG is not a one-size-fits-all solution, as the GPT-4 model equipped with the ReAct falls short of generating comprehensive content. This is attributed to the fact that ReAct repurposes the GPT into a role more akin to planner and executor, constraining its capacity as a parametric knowledge base.

\begin{figure}[t]
  \centering
  \includegraphics[scale = 0.6]{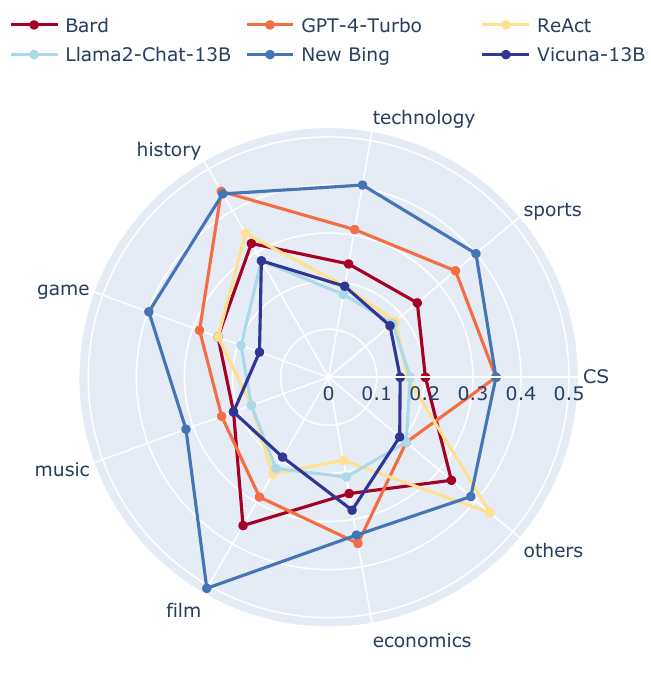}
  \caption{Performance of LLMs on different domains.}
  \label{fig:domain}
\end{figure}

\subsection{Domain}
\label{sec:domain_r}
\textit{Proprietary LLMs overwhelmingly outperform other competitors in all domains. The advantage is further extended by integrating the search engine.}
Figure~\ref{fig:domain} shows that the GPT-4-Turbo surpasses the open-sourced LLMs in all aspects. However, it is worth noting that in some domains, such as music and economics, the gap between open-source models and GPT-4-Turbo is very small, but open-sourced LLMs are biased and inadequate to cover all domains. Training LLMs that excel at multiple domains remains sufficient exploration.

\subsection{Impact of Alternate Annotation}
\label{sec:alter_anno}
\begin{figure}[t]
    \centering
    \includegraphics[width=\linewidth]{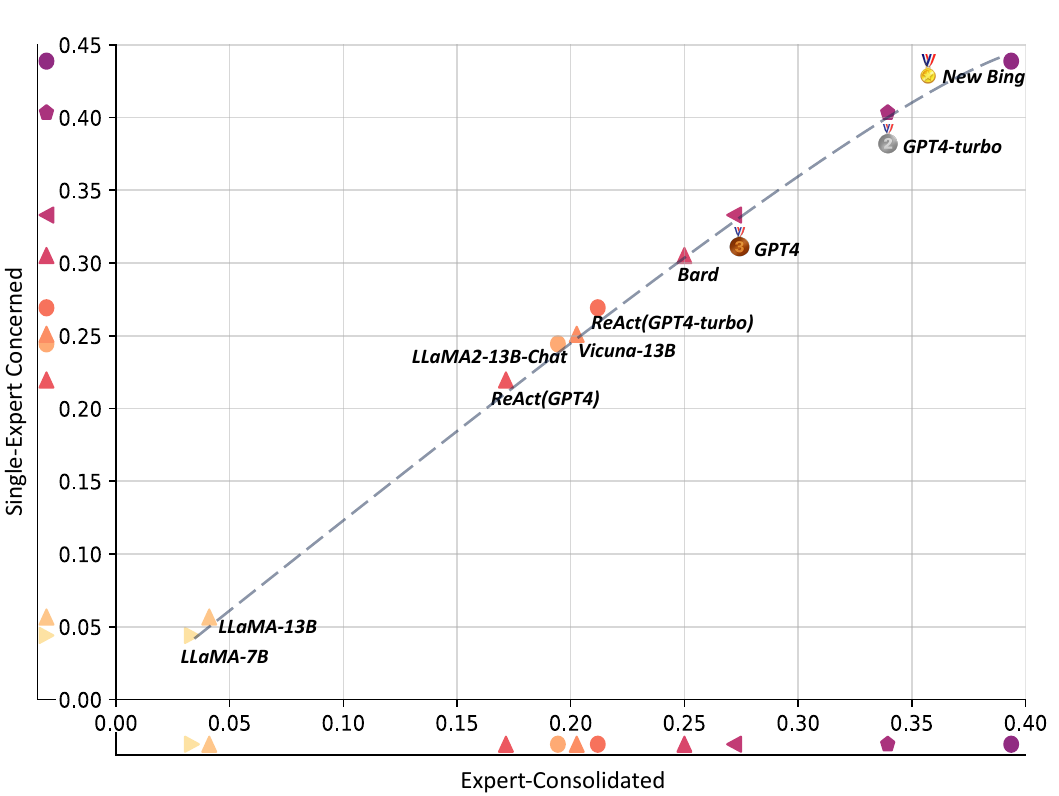}
    \caption{Performance difference on the experts-consolidated and single-expert-focused set.}
    \label{fig:multi-round}
\end{figure}

\textit{Despite the impressive performance, proprietary LLMs are still unable to cater to the preferences of every individual. }
Figure~\ref{fig:multi-round} compares the LLMs' performance on expert-consolidated and single-expert-focused subsets. As the iterative verification and supplementation of proxy-questions proceed, the performance of all models decreases, suggesting that LLMs cannot cater to every individual's preferences. Remarkably, New Bing outperforms all other baselines by a considerable margin, no matter the sub-split where only a single expert is involved or on the complete expert-consolidated set. However, despite the impressive performance, significant degradation could be observed. This suggests that the multi-perspective evaluation criteria in \textsc{ProxyQA} pose critical challenges to LLMs.

\subsection{Generation Length}
\label{sec:gen_len}
\textit{Improving the readability and informativeness of generated content within limited token budgets remains an area for systematic exploration.}
The average generation lengths are presented in Table~\ref{tab:avg_len}. It is essential to emphasize that the degree of informativeness and comprehensiveness is not proportional to the length of the generated content. Specifically, LLaMA2-13B generates lengthy content, yet it exhibits the lowest quality in generating contextual background on \textsc{ProxyQA}. In contrast, GPT-4-Turbo produces concise content while conveying extensive and comprehensive information. Moreover, when GPT-4 is incorporated with the search engine, the New Bing Creative Mode yields highly informative and in-depth content, significantly surpassing all other baseline models with an acceptable increase in generation length. 

\input{tables/avg_len}

%% file: tables/avg_len.tex
\begin{table}[t]
\centering

\scalebox{0.83}
{
\begin{tabular}{lcc}
\toprule
                   & \textbf{Avg. Len.} & \textbf{ Acc.} \\ \hline
LLaMA2-13B         & 1906.87            & 4.11                       \\
LLaMA2-13B-Chat    & 869.42             & 19.44                      \\
Vicuna-13B         & 727.84             & 20.26                      \\
\midrule
GPT-3.5-Turbo      & 823.32             & 23.94                      \\
GPT-4              & 744.00              & 27.19                      \\
GPT-4-Turbo        & 1029.47            & 33.94                      \\
\midrule
ReAct (GPT-4-Turbo) & 355.80             & 21.19                      \\
Bard (Gemini Pro)   & 922.83             & 25.00                      \\
New Bing (Creative)          & 1167.65            & \textbf{39.37}                      \\ \bottomrule
\end{tabular}
}
\caption{Average word count of the generated reports.}
\label{tab:avg_len}
\end{table}

%% file: sections/analysis.tex
\section{Analysis}
\subsection{Win Rate}
\label{sec:win_rate}
We study the pairwise win rate among various LLMs evaluated by \textsc{ProxyQA} and compare the results with human evaluation to validate the effectiveness of \textsc{ProxyQA}. 

\paragraph{Setup}
Five well-educated postgraduate students are engaged; all have not participated in annotating the meta and proxy-questions of \textsc{ProxyQA}; they are required to score and rank the randomly sampled reports generated by different LLMs. The scoring guideline is shown in Appendix~\ref{appendix:score_guide}. We sampled ten meta-questions from \textsc{ProxyQA} and employed four LLMs to generate comprehensive reports. As a further comparison, we also follow the settings in MT-Bench~\citep{zheng2023judging} that adopt LLM-as-judges, which directly rate the generated report based on the scoring guideline. Similarly, we utilize GPT-Seperate (GPT-S), which evaluates a single report at a time, and GPT-Batch (GPT-B), which evaluates and compares multiple reports simultaneously, to score and rank each report. Given that five human evaluators are involved in the comparison, we ensure fairness and robustness by requiring GPT-S, GPT-B, and our proposed \textsc{ProxyQA} to evaluate each report five times. We then calculate the average win rate based on the pairwise comparison. 

\begin{figure}[t]
    \raggedleft
    \includegraphics[width=\linewidth]{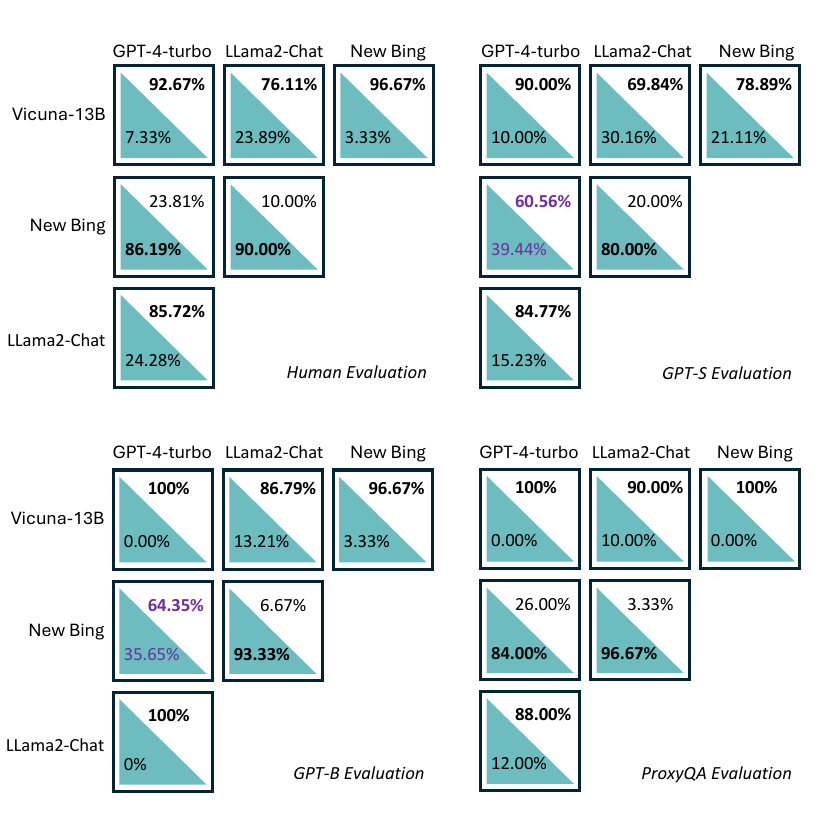}
    \caption{Comparison of win rate of various evaluation methods. GPT-evaluators are highly overconfident in the results produced by GPT-4-Turbo, while \textsc{ProxyQA} significantly correlated with human preference.}
    \label{fig:win_rate}
    \vspace{-1em}
\end{figure}

\paragraph{Result}\textit{GPT-as-judges over-confident on the contents generated by GPT models, while ProxyQA's choice is highly correlated with humans.} As shown in Figure~\ref{fig:win_rate}, majority of the evaluation results highly correlated with the human's choices. Specifically, evaluators generally recognize the quality of the reports generated by GPT-4-Turbo and New Bing, i.e., their win rates are much higher than those of Vicuna and LLama2-Chat. It is worth noting, however, that GPT-S and GPT-B exhibit overconfidence in the quality of the reports generated by GPT-4-Turbo compared to New Bing. In contrast, both human evaluators and ProxyQA exhibit a preference for New Bing over GPT-4-Turbo. This outcome attests to the effectiveness of ProxyQA and demonstrates the correlation between ProxyQA and human evaluation.

\subsection{Agreement Evaluation} 
 To thoroughly examine the consistency of \textsc{ProxyQA} and the correlation against human evaluation criteria, we investigate the agreement rate of human assessment and \textsc{ProxyQA}. Two categories of agreement rates are explored. Similar to MT-Bench~\cite{zheng2023judging}, to effectively gauge the consistency of our proposed \textsc{ProxyQA}, we assess the self-agreement rate, which calculates the inter-evaluator agreement rate. Furthermore, we establish that the evaluation method proposed in the \textsc{ProxyQA} exhibits a strong correlation with human judgment by determining the agreement rate between human evaluations, referred to as the human agreement rate. We employ GPT-Seperate and GPT-Batch, as discussed in section~\ref{sec:win_rate}, for comparison purposes.

\paragraph{Setup} Following the settings in section~\ref{sec:win_rate}, 20 reports generated by GPT-4-Turbo and New Bing are evaluated and compared by experts that have not participated in the annotation of \textsc{ProxyQA}. We analyze the agreement between different evaluation methods and human evaluations.

\paragraph{Self-agreement}
Given a pair of meta-question and its corresponding generated report, each evaluation method is required to score and vote the preferred reports $n$ times. Let $V = \{v_1, \dots, v_n\}$ be the set of voting results, then the self-agreement rate is calculated as:

\begin{equation}
    R_{self}  \triangleq \frac{1}{C_n^{2}}\sum_{i=1}^n\sum_{j=i+1}^n \mathds{1} \{v_i = v_j\}
\end{equation}

Where $\mathds{1}\{\cdot\}$ denote the indicator function. The self-agreement rate, denoted as $R_{self}$, quantifies the consistency of an evaluation method.

\paragraph{Human-agreement} The calculation is divided into Majority-to-Group (M2G), Group-to-Majority (G2M) and Group-to-Group (G2G). 
The $R_{M2G}$ quantifies the proportion of the majority vote of a specific evaluation method in concordance with the overall votes of human evaluation. 
Conversely, $R_{G2M}$ calculates the proportion of overall votes of an evaluation method that concur with the majority vote of humans, indicating how well the evaluation criteria are aligned with the majority opinion of humans.
$R_{G2G}$ provides a view of the overall agreement between an evaluation method and the human.
Let $V_e = \{e_1, \dots, e_n\}, V_h = \{h_1, \dots, h_n\}$ represent the set of the voting results of an evaluation method and human, respectively. The majority vote of a set is represented as $\mathds{M}(\cdot)$. Then, the agreement is calculated as:
\begin{align}
    R_{M2G} \triangleq \frac{1}{n}\sum_{i=1}^n \mathds{1} \{h_i = \mathds{M}(V_e)\} \\
    R_{G2M} \triangleq \frac{1}{n}\sum_{i=1}^n \mathds{1} \{e_i = \mathds{M}(V_h)\} \\
    R_{G2G} \triangleq \frac{1}{n^2} \sum_{i=1}^n\sum_{j=1}^n \mathds{1} \{e_i = h_j\} 
\end{align}

\input{tables/agreement} 

\paragraph{Results}
\textit{Human preference varies from individual to individual, while ProxyQA shows firm consistency and is highly correlated to the majority of humans.}
Table~\ref{tab:agree} illustrates that \textsc{ProxyQA} offers a highly consistent metric for evaluating long content generation. Furthermore, \textsc{ProxyQA} strongly correlates with the majority opinion of human evaluators, thereby emphasizing its efficacy in validating long-form content. Notably, the subjective preferences of human experts vary, as the agreement rate reaches only 52.19\%, leading to relatively inconsistent evaluations, which is consistent with the findings of \citet{XuSIC23}. In contrast, \textsc{ProxyQA} achieves an 88.00\% agreement rate, indicating its potential as a highly consistent performance indicator. Moreover, when evaluating the consensus between different evaluation methods and the majority opinion of human experts, both GPT-S and GPT-B are contrary to human preference. GPT-as-judges is overconfident in the reports generated by GPT-4-Turbo, which becomes even more extreme when evaluating with GPT-B. However, the criteria of \textsc{ProxyQA} significantly align with the majority opinion of human experts, surpassing GPT-as-judges in all human-agreement rates with a substantial margin and attaining 66.19\% and 63.33\% M2G and G2M agreement rate respectively. These findings provide robust evidence that the proposed \textsc{ProxyQA} can effectively and reliably assess the capabilities of LLMs in generating long-form content.

\subsection{Validation of the GPT Evaluator}
To ensure the evaluator's accuracy and confirm that answers to proxy-questions are derived strictly from the information within the generated content, we involved five human experts to conduct a human evaluation. If the GPT-evaluator fails to extract correct information from the generated content to answer a proxy-question, or answers a proxy-question without using the knowledge presented in the content (answer using the internal knowledge), the experts will consider it a negative sample. We validated the accuracy of our proxy-evaluator. Given the reports generated by GPT-4-Turbo, New Bing, Vicuna-13B, and LLama2-Chat, 100 proxy-questions and the boolean answer generated by the GPT-4 evaluator are sampled. The GPT-4 evaluator reaches a 91\% accuracy rate, demonstrating its capability as a reliable evaluator.

\subsection{Validation of Prompts}
Furthermore, we evaluated the model's performance using different prompts such as ``Please write a comprehensive long report on the following question''. We compared the results to the performance without any prompts. Our analysis indicated that, in most instances, the presence or absence of different prompts had minimal impact on performance. We attribute this to the fact that the majority of meta-questions in ProxyQA already necessitate the LLM to ``explain/introduce/compare in detail'' or ``generate a detailed report about .''

%% file: tables/agreement.tex
\begin{table}[t]
\centering
\scalebox{0.85}
{
\begin{tabular}{lcccc}
\toprule
        & Self           & G2G            & G2M            & M2G            \\ \hline
GPT-S   & 48.65          & 45.49          & 47.62          & 30.00          \\
GPT-B   & 51.17          & 43.86          & 36.66          & 36.66          \\
ProxyQA & \textbf{88.00} & \textbf{66.00} & \textbf{63.33} & \textbf{66.19} \\
Human   & 52.19          & -              & -              & -              \\ 
\bottomrule
\end{tabular}
}
\caption{Agreement between each evaluation method and human evaluation. Self-agreement is also reported.}
\label{tab:agree}
\end{table}

%% file: sections/conclusion.tex
\section{Conclusion}

In this work, we introduce \textsc{ProxyQA}, a framework designed to evaluate LLMs' ability to generate long-form text. Unlike traditional methods that rely on a direct comparison with a reference text, by employing an evaluator to use the information provided in the LLM-generated text to answer proxy-questions, the framework assesses the LLMs' ability without fixed references or crowdsource workers. By mitigating concerns over data contamination and ensuring the relevance and freshness of evaluation content, \textsc{ProxyQA} enhances our understanding of LLMs and drives innovation towards developing long-form generation methods with LLMs.

%% file: sections/limitation.tex
\section{Limitation}
In this study, \textsc{ProxyQA} mainly focuses on evaluating the informative and knowledge coverage of the generated texts. However, multiple key dimensions such as factuality~\citep{MinKLLYKIZH23}, verifiability~\citep{LiuIXWXZ23}, and coherency~\citep{DengKR22} should be considered for long-form content generation. For instance, proxy questions cannot measure hallucination in long-form content, a critical issue for a long-form generation. On the other hand, each meta-question is annotated by five experts, but they cannot cover potentially all the proxy-questions. More advanced methods that consider these issues will be developed in future work. In addition, due to limited annotation resources, we prioritized adding as many proxy-questions as possible to maintain a high benchmark quality, rather than increasing the number of meta-questions. More meta-questions should be included in the future version.

\section*{Ethical Considerations} \label{apx:ethical}
To avoid potential ethical issues, we carefully checked all questions in multiple aspects, as discussed in Section~\ref{sec:quality}. We try to guarantee that all samples do not involve any offensive, gender-biased, or political content, and any other ethical issues. The source code will be released with instructions to support correct use. The baseline model we tested are all open-sourced LLMs and public APIs, these previous works have already considered ethical issues when creating the models.

\section*{Acknowledgements}
We would like to express our profound gratitude to the anonymous reviewers for their invaluable and insightful feedback. This research has been partially funded by the Research Grants Council of the Hong Kong SAR under Grant GRF 11217823 and the Collaborative Research Fund C1042-23GF. Additional support was provided by the National Natural Science Foundation of China under Grant 62371411, the InnoHK initiative, the Government of the HKSAR, and the Laboratory for AI-Powered Financial Technologies.

%% file: sections/appendix.tex
\section{Appendix}
\label{sec:appendix}

\subsection{Advancements in Long-Context LLMs}

As LLMs gain global traction, their ability to adeptly process extensive sequences—such as protracted conversation histories or intricate scientific documents—becomes increasingly significant. Recent years have witnessed the emergence of LLMs capable of managing long context windows, a feat made possible by the advent of more powerful GPUs with expanded memory capacities, along with innovations in memory-efficient attention mechanisms~\citep{DaoFERR22, Dao2023}. Historically, the context window size for models like GPT-2 encompassed 1024 tokens~\citep{Radford2019LanguageMA}, which was then extended to 2048 in GPT-3~\citep{BrownMRSKDNSSAA20}. Modern iterations, such as GPT-4-turbo, boast an impressive 128K token capacity~\citep{gpt4128k}, while Claude 2.1 extends this even further to 200K tokens~\citep{claude200k}. Nevertheless, scaling the context window during the pretraining phase remains a daunting task as the computational demands surge quadratically with the length of the attention span, and a majority of texts within standard corpora, like Common Crawl, tend to be comparatively brief.

A novel approach gaining momentum among researchers is the augmentation of the LLMs' context window through the process of continued training or fine-tuning. For instance, \citet{Tworkowski2023} successfully refined the 3B and 7B OpenLLaMA checkpoints, employing contrastive training techniques to adeptly handle contexts stretching up to 8K tokens. Similarly, \citet{Mohtashami2023} achieved an expansion of the context length from 4K to 32K for LLaMA 7B by incorporating ``landmark tokens'' that effectively encapsulate blocks of the existing context. These tokens allow for focused fine-tuning of attention mechanisms, which in turn facilitates the selection of pertinent contextual blocks.

Furthermore, \citet{Chen2023} introduced a method known as positional interpolation, to be used with LLMs that incorporate Rotary Position Embeddings (RoPE) as their choice of positional encoding~\citep{Su2021RoFormerET}. This technique yielded promising outcomes when applied to LLaMA models ranging from 7B to 65B in size, requiring minimal fine-tuning efforts—a mere 1000 optimization steps. A different paradigm, ALiBi \citep{PressSL22}, circumvents the necessity of fine-tuning altogether for expanding the context window. By eschewing positional embeddings and instead applying a linear bias to the attention scores—which is proportionate to the distance between tokens—it elegantly adjusts to handle longer contexts.

Lastly, the strategy proposed by~\citet{Ratner2022ParallelCW} partitions extensive contexts into several sub-windows, employing the same positional embeddings across them. This innovative reuse of embeddings enables the models to cope with extended contexts without the need for additional fine-tuning. This collective body of work represents the ongoing evolution of strategies to enhance the capabilities of LLMs in accommodating long context sequences, a critical requirement for their effective deployment in complex, real-world applications.

\subsection{Evaluation for Long-Context LLMs}
The advent of long-context LLMs has ushered in an era where evaluating performance over extensive text sequences is crucial. Benchmarks like ZeroSCROLLS~\citep{shaham2023} have emerged to challenge these models' understanding of expansive texts in a zero-shot setting. ZeroSCROLLS extends the foundation laid by the SCROLLS benchmark~\citep{shaham2022}—originally designed to handle longer texts through fine-tuning—by introducing four new tasks: query-based summarization, multi-hop question answering, sentiment aggregation, and ordering book chapter summaries. It distinguishes itself by focusing on zero-shot performance, using simple natural language prompts and eschewing training data, relying on non-public, high-quality references.

Another contribution to this domain is LongBench~\citep{Bai2023}, a suite of 21 datasets across 6 categories of tasks such as single- and multi-document question answering, summarization, few-shot learning, specific synthetic tasks, and code completion. What sets LongBench apart is its uniform format for all datasets, promoting a unified and automated evaluation process with metrics like F1 and ROUGE. Bamboo~\citep{Dong2023} also provides a valuable framework for analyzing comprehension over lengthy texts, offering a selection of 10 datasets from 5 diverse activities that range from question answering to hallucination detection, text sorting, language modeling, and code completion. Bamboo specifically tackles potential data contamination by exclusively using sources released no earlier than 2023, maintaining the relevance and contemporaneity of its material.

L-Eval~\citep{An2023} introduces a bifurcated approach to evaluate LLMs, featuring both closed-ended and open-ended tasks. Closed-ended tasks focus on the model's reasoning and comprehension skills in a protracted context. In contrast, its open-ended tasks provide a variety of summarization challenges that require models to synthesize information from lengthier documents. InfiniteBench~\cite{zhang2023infinitebench} is tailored to assess LLMs that process, understand, and infer information from contexts that span over 100,000 tokens. It should be noted that these datasets prioritize the assessment of long-context understanding, and as a result, a significant portion of tokens are used as inputs for the LLMs rather than outputs.

\subsection{Baseline Models}
\label{appendix:baseline}
All the baselines are prompted with ``Write a well-structured and extensive report to answer the question: [META QUESTION]''. We then employ the generated results as the contextual background and force the GPT-4 and GPT-4-turbo to answer the proxy-questions accordingly.
Multiple competitive baselines are tested under \textsc{ProxyQA}.

\paragraph{Base LLaMA} is a set of open-sourced LLMs pretrained on diverse sources spanning multiple domains~\cite{touvron2023, llama2}. The pretraining corpora include the June 2022 Wikipedia dumps, which should enable the LLaMA family to effectively address most `easy' meta-questions in \textsc{ProxyQA}. Our experiment evaluates LLaMA-7B, LLaMA2-7B, and LLaMA2-13B. 

\paragraph{Instruction-Finetuned LLaMA} includes Vicuna~\cite{vicuna2023}, Alpaca~\cite{alpaca}, LLaMA2-Chat~\cite{llama2}. Vicuna is a chat assistant trained by fine-tuning LLaMA on around 70k user-shared conversations collected from ShareGPT~\cite{sharegpt}. Similarly, Alpaca is trained with 52k self-instructed demonstrations adapted text-davinci-003~\cite{textdavinci}. As an extension of base LLaMA2, LLaMA2-Chat is optimized specifically for dialogue usage of over 1 million instructions. 

\paragraph{OpenAI APIs} includes GPT-3.5-turbo\cite{gpt35turbo}, GPT-4 and GPT-4-turbo~\cite{gpt4}. The default decoding configuration is utilized to generate responses, while the maximum decoding length is set as its maximum limitation. Both GPT-3.5-turbo and GPT-4-turbo are of version 1106, while GPT-4 corresponds to GPT-4-0613. The training data for GPT-4-turbo is up-to-date as of April 2023, while the remaining models are trained with data up to September 2021.

\paragraph{Web-Augmented LLMs} utilize external search APIs are evaluated. Specifically, GPT-4 and GPT-4-turbo are integrated with the Google Search API under the configuration of ReAct~\cite{yao2023react}. These models are tasked with processing meta-questions, reasoning through search traces, and extracting relevant content from search results from the internet across multiple turns. The implementation is adopted from LangChain~\cite{langchain}. In addition to ReAct, the performance of New Bing (creative mode)~\cite{new_bing} and Bard with Gemini Pro~\cite{geminiteam2023gemini} are also assessed. 

\subsection{Scoring Guideline for Human Evaluation and LLM-as-Judges}
\label{appendix:score_guide}
\begin{tcolorbox}[title=Scoring Guideline]
Please rate the knowledge coverage of the reports provided, using a scale of 0-5. Assess how well the report covers the necessary information related to the question. \\
Knowledge Coverage Scale: \\
0 - Nonsense: The report offers no useful information and is completely irrelevant to the question. \\
1 - Poor: The report provides very little useful information and barely addresses the question. \\
2 - Fair: The report offers some useful information but lacks depth and detail, leaving the question partially unanswered. \\
3 - Average: The report presents a decent amount of information, addressing the question adequately but not exceptionally. \\
4 - Good: The report provides comprehensive information, covering the question well with appropriate depth and detail. \\
5 - Excellent: The report thoroughly covers all aspects of the question, offering a high level of detail and leaving no gaps in knowledge.
\end{tcolorbox}

\subsection{Prompts for tested LLMs and Evaluator}
\label{appendix:prompt}
All the tested baselines are prompted with ``Write a well-structured and extensive report to answer the question: [META QUESTION]'', while the evaluators are prompted with:
\begin{tcolorbox}[title=Prompt for Evaluator]
    Read the provided document and determine whether the statement below is "True" or "False". Use only the information in the text to make your decision. Do not rely on prior knowledge or information outside of the given text. If the text does not provide enough information to make a decision, respond with "Not mentioned". \\
    Format your answer as "True", "False", or "Not mentioned". \\
    Document: [generated\_report] \\
    Statement: [proxy\_question]
\end{tcolorbox}

\subsection{Annotation Guideline for Formulating Meta-Questions and Proxy-questions}
\begin{tcolorbox}
The meta-questions should be based on the topic the experts are most familiar with or keen on.

Meta-question: The meta-question should be: \\
a) Answerable through thorough research. 
b) Aligned with real-life scenarios. 
c) Avoid of offensive or ethical concerns. 
d) Without an absolute or unique answer. 
e) Addressable in long-form reports or articles. 
f) Ensure the questions are open-ended, promoting in-depth research and discussion. 

Proxy-questions: For each meta-question, determine the essential proxy-questions that cover the critical contents of the topic. proxy-questions should be: \\
a) Directly related to the meta-question. 
b) Comprehensive enough to cover different angles of the meta-question. 

Golden Answer: For each proxy-question, label the golden answer, which refers to the most accurate and relevant information. 

Steps for Annotation: \\
a) Choose a topic you are familiar with or interested in. 
b) Formulate a meta-question following the criteria mentioned above 
c) Determine the essential proxy-questions that cover the critical contents of the topic, as per the guidelines above. 
d) For each proxy-question, label the golden answer. 
e) Review the meta-question, proxy-questions, and golden answer to ensure they adhere to the guidelines and criteria provided. 
f) Collaborate with your peers to review and refine the meta-questions and proxy-questions. Verify and supplement the proxy-questions in an alternate way. 

\end{tcolorbox}

\subsection{Case Study and Error Analysis}
Table~\ref{tab:case}-~\ref{tab:case2-gpt4t} show the case study of the reports generated by LLaMA2-13B-Chat, New Bing and GPT-4-turbo. Although LLaMA2-13B-Chat generates lengthy contents, very little important information are contained. Therefore GPT-4 evaluator is unable to fetch useful information from the generated contents, result in low accuracy on the proxy-questions. In contrast, both New Bing and GPT-4-turbo generate more concise results conveyed with essential contents. Specifically, critical hallucination can be observed in the output of the LLaMA2-13B-Chat, while none of the contents can be used to answer the proxy-questions. However, New Bing covers as much points as it can, providing a comprehensive overview of the given topic, while results of GPT-4-turbo is well-structured and offers detailed and helpful introduction to some of the critical contents.

\subsection{Distribution of the ProxyQA}
\label{appendix:dist}
51\% of meta-questions are categorized as easy, signifying that they can be addressed utilizing information solely from Wikipedia. In contrast, 49\% of meta-questions are deemed difficult. Moreover, 53\% of these questions are time-sensitive, as their answers may dynamically alter over time. A representative example of a time-sensitive question is: "Could you kindly offer a comprehensive overview of the historical development of pre-training in NLP, encompassing notable contributions?"

\subsection{Maintaince of the Leaderboard}
\label{appendix:mt}
Concerning meta-questions of varying difficulty levels, the average accuracy of evaluators will be reported for both easy and hard subsets of \textsc{ProxyQA}. In addition, for time-sensitive questions, proxy-questions will be incrementally updated, and the leaderboard will be dynamically adjusted to evaluate the LLMs' possession of current knowledge. The leaderboard is updated on a weekly basis, while the proxy-question of time-sensitive meta-question is incrementally updated on a bi-monthly basis. Each periodic update of proxy-questions will also renew the leaderboard.

\begin{table*}[!t]
\centering
\resizebox{\textwidth}{!}{
\begin{tabular}{p{15cm}}
\textbf{Question:} Data parallelism, model parallelism, and pipeline parallelism play a vital role in the training of large-scale language models. What are the representative works and frameworks among these technologies? Please introduce these technologies and frameworks in detail. \\
\\
\toprule
\multicolumn{1}{c}{LLaMA2-13B-Chat-Page 1}  \\ \hline
\textbf{Introduction:} \\
The training of large-scale language models has become increasingly popular in recent years due to their wide range of applications in natural language processing (NLP) tasks. However, training such models can be computationally expensive and time-consuming, especially when dealing with large datasets. To address this challenge, researchers have proposed various parallelization techniques to speed up the training process. In this report, we will focus on three key parallelization techniques: data parallelism, model parallelism, and pipeline parallelism. We will introduce representative works and frameworks that utilize these techniques and discuss their advantages and limitations. \\
\textbf{Data Parallelism:} \\
Data parallelism is a technique that involves dividing the input data into smaller chunks and processing them in parallel across multiple devices or machines. This technique is particularly useful for training large-scale language models, as it allows for faster training times and more efficient use of computational resources. \\
One of the most popular frameworks for data parallelism is TensorFlow, which is an open-source machine learning library developed by Google. TensorFlow provides a built-in data parallelism feature called TensorFlow Estimator, which allows users to parallelize their training processes across multiple machines. Another popular framework for data parallelism is PyTorch, which provides a module called PyTorch Distributed that allows users to parallelize their training processes across multiple devices. \\
\textbf{Representative Works:} \\
One of the most influential works in data parallelism for large-scale language models is the paper "Distributed Training of Deep Neural Networks" by Google researchers (2016). This paper introduced the concept of using data parallelism to train deep neural networks on large datasets, and demonstrated the effectiveness of this approach for training large-scale language models. \\
Another influential work is the paper "Deep Learning with CUDA: General Purpose Training of Neural Networks on GPU Clusters" by NVIDIA researchers (2017). This paper introduced the concept of using GPU clusters for training large-scale language models, and demonstrated the effectiveness of this approach for improving training times and reducing computational costs. \\
\textbf{Model Parallelism:} \\
Model parallelism is a technique that involves dividing the model into smaller parts and processing them in parallel across multiple devices or machines. This technique is particularly useful for training large-scale language models, as it allows for faster training times and more efficient use of computational resources. \\
One of the most popular frameworks for model parallelism is the open-source machine learning library TensorFlow, which provides a built-in feature called TensorFlow Model Parallelism. This feature allows users to parallelize their training processes across multiple machines, and has been used in a number of influential works for training large-scale language models. \\
Another popular framework for model parallelism is the PyTorch framework, which provides a module called PyTorch Model Parallel that allows users to parallelize their training processes across multiple devices. \\
\bottomrule
\end{tabular}
}
\caption{Case study of LLaMA2-13B-Chat}
\label{tab:case}
\end{table*}

\begin{table*}[!t]
\centering
\resizebox{\textwidth}{!}{
\begin{tabular}{p{15cm}}
\toprule
\multicolumn{1}{c}{LLaMA2-13B-Chat-Page 2}  \\ \hline
\textbf{Representative Works:} \\
One of the most influential works in model parallelism for large-scale language models is the paper "Model Parallelism for Training Large-Scale Neural Networks" by Google researchers (2017). This paper introduced the concept of using model parallelism to train large-scale language models, and demonstrated the effectiveness of this approach for improving training times and reducing computational costs. \\
\textbf{Pipeline Parallelism:} \\
Another influential work is the paper "Scaling Up Neural Machine Translation with Model Parallelism" by Google researchers (2018). This paper introduced the concept of using model parallelism to train large-scale neural machine translation models, and demonstrated the effectiveness of this approach for improving translation quality and reducing computational costs. \\
Pipeline parallelism is a technique that involves dividing the training process into smaller stages and processing them in parallel across multiple devices or machines. This technique is particularly useful for training large-scale language models, as it allows for faster training times and more efficient use of computational resources. \\
One of the most popular frameworks for pipeline parallelism is the open-source machine learning library TensorFlow, which provides a built-in feature called TensorFlow Pipeline Parallelism. This feature allows users to parallelize their training processes across multiple machines, and has been used in a number of influential works for training large-scale language models. \\
Another popular framework for pipeline parallelism is the PyTorch framework, which provides a module called PyTorch Pipeline that allows users to parallelize their training processes across multiple devices. \\
\textbf{Representative Works:} \\ 
One of the most influential works in pipeline parallelism for large-scale language models is the paper "Pipeline Parallelism for Training Large-Scale Neural Networks" by Google researchers (2018). This paper introduced the concept of using pipeline parallelism to train large-scale language models, and demonstrated the effectiveness of this approach for improving training times and reducing computational costs. \\
Another influential work is the paper "Scaling Up Neural Machine Translation with Pipeline Parallelism" by Google researchers (2019). This paper introduced the concept of using pipeline parallelism to train large-scale neural machine translation models, and demonstrated the effectiveness of this approach for improving translation quality and reducing computational costs. \\
\textbf{Conclusion: }\\
In conclusion, data parallelism, model parallelism, and pipeline parallelism are three key parallelization techniques that have played a vital role in the training of large-scale language models. These techniques have been used in a number of influential works, and have demonstrated the effectiveness of parallelizing the training process to improve training times and reduce computational costs. As the field of NLP continues to evolve, it is likely that these parallelization techniques will continue to play an important role in the training of large-scale language models. \\

\bottomrule

\end{tabular}
}
\caption{Case study of LLaMA2-13B-Chat}
\label{tab:case1}
\end{table*}

\begin{table*}[!t]
\centering
\resizebox{\textwidth}{!}{
\begin{tabular}{p{15cm}}
\toprule
\multicolumn{1}{c}{New Bing - Page 1}  \\ \hline
\input{tables/new_bing_case1} \\
\bottomrule
\end{tabular}
}
\caption{Case study of New Bing}
\label{tab:case-bing}
\end{table*}

\begin{table*}[!t]
\centering
\resizebox{\textwidth}{!}{
\begin{tabular}{p{15cm}}
\toprule
\multicolumn{1}{c}{New Bing - Page 2}  \\ \hline
\input{tables/new_bing_case2} \\
\bottomrule
\end{tabular}
}
\caption{Case study of New Bing}
\label{tab:case-bing2}
\end{table*}

\begin{table*}[!t]
\centering
\resizebox{\textwidth}{!}{
\begin{tabular}{p{15cm}}
\toprule
\multicolumn{1}{c}{GPT-4-turbo - Page 1}  \\ \hline
\input{tables/gpt4t-case} \\
\bottomrule
\end{tabular}
}
\caption{Case study of GPT-4-turbo}
\label{tab:case-gpt4t}
\end{table*}

\begin{table*}[!t]
\centering
\resizebox{\textwidth}{!}{
\begin{tabular}{p{15cm}}
\toprule
\multicolumn{1}{c}{GPT-4-turbo - Page 2}  \\ \hline
\input{tables/gpt4t-case2} \\
\bottomrule
\end{tabular}
}
\caption{Case study of GPT-4-turbo}
\label{tab:case2-gpt4t}
\end{table*}

%% file: tables/new_bing_case1.tex
Hello, this is Bing. I can help you with your question.

Data parallelism, model parallelism, and pipeline parallelism are three
types of parallelism methods that can be used to train large-scale
language models efficiently on multiple GPUs or other devices. Here is a
brief introduction of these technologies and some representative works
and frameworks that use them:

\begin{itemize}
\item
  \textbf{Data parallelism (DP)}: This is the simplest and most common
  form of parallelism, where the same model is replicated on multiple
  devices, and each device processes a different slice of the input
  data. The gradients are then aggregated and synchronized across the
  devices to update the model parameters. This method is easy to
  implement and can scale well with the number of devices, but it has
  some limitations, such as requiring large memory to store the full
  model and communication overhead for gradient synchronization. Some
  examples of frameworks that support data parallelism are PyTorch's
  \textbf{\href{\%5E1\%5E}{DistributedDataParallel}} and TensorFlow's
  \textbf{\href{\%5E6\%5E}{MirroredStrategy}}.
\item
  \textbf{Model parallelism (MP)}: This is a more complex form of
  parallelism, where the model is split into smaller parts and
  distributed across multiple devices, such that each device only stores
  and computes a subset of the model parameters. This method can
  overcome the memory limitation of data parallelism and enable training
  very large models that cannot fit on a single device. However, it also
  introduces more challenges, such as how to partition the model, how to
  minimize the communication cost, and how to balance the workload among
  the devices. There are different types of model parallelism, such as
  tensor parallelism and pipeline parallelism, which will be explained
  below. Some examples of frameworks that support model parallelism are
  NVIDIA's \textbf{\href{\%5E2\%5E}{Megatron-LM}} and Microsoft's
  \textbf{DeepSpeed}.
\item
  \textbf{Tensor parallelism (TP)}: This is a type of model parallelism
  where each tensor (such as a weight matrix or an activation vector) is
  split into smaller chunks along one or more dimensions, and each chunk
  is assigned to a different device. For example, a weight matrix of
  size {[}m, n{]} can be split into {[}m/k, n/k{]} chunks, where k is
  the number of devices, and each device only stores and computes one
  chunk. This method can reduce the memory footprint and communication
  cost by a factor of k, but it also requires modifying the model
  architecture and the computation logic to handle the tensor sharding.
  Some examples of works that use tensor parallelism are \textbf{gShard}
  and \textbf{Mesh-TensorFlow}.
\item
  \textbf{Pipeline parallelism (PP)}: This is another type of model
  parallelism where the model is split into smaller stages (such as
  layers or groups of layers) and distributed across multiple devices,
  such that each device only stores and computes one stage of the model.
  For example, a model with L layers can be split into S stages, where
  each stage contains L/S layers, and each device only processes one
  stage. The input data is then fed into the first device, and the
  output of each device is passed to the next device in a pipelined
  fashion. This method can also reduce the memory footprint and
  communication cost by a factor of S, but it also requires careful
  scheduling and synchronization of the pipeline stages to maximize the
  throughput and avoid idle periods. Some examples of works that use
  pipeline parallelism are \textbf{PipeDream}, \textbf{GPipe}, and
  \textbf{\href{\%5E4\%5E}{PipeEdge}}.
\end{itemize}

%% file: tables/new_bing_case2.tex
These parallelism methods can be combined in different ways to achieve
higher efficiency and scalability. For example, one can use data
parallelism and tensor parallelism together to train a large model on a
large number of devices, as done in
\textbf{\href{\%5E2\%5E}{Megatron-LM}}. Alternatively, one can use data
parallelism and pipeline parallelism together to train a deep model with
high throughput, as done in \textbf{GPipe}. Furthermore, one can use
tensor parallelism and pipeline parallelism together to train a very
large and deep model with low memory and communication cost, as done in
\textbf{DeepSpeed}.

I hope this helps you understand the basics of parallelism methods for
large-scale language model training. If you want to learn more, you can
check out the links I provided or search for more resources on Bing.
Have a nice day!

\begin{enumerate}
\def\labelenumi{(\arabic{enumi})}
\item
  Model Parallelism - Hugging Face.
  https://huggingface.co/docs/transformers/v4.15.0/parallelism.
\item
  undefined. https://github.com/nvidia/megatron-lm.
\item
  Efficient Large-Scale Language Model Training on GPU Clusters -
  arXiv.org. https://arxiv.org/pdf/2104.04473v2.
\item
  PipeEdge: Pipeline Parallelism for Large-Scale Model Inference on
  \ldots. https://ieeexplore.ieee.org/document/9996638.
\item
  Efficient Large-Scale Language Model Training on GPU Clusters -
  arXiv.org. https://arxiv.org/pdf/2104.04473v2.
\item
  Optimizing Distributed Training on Frontier for Large Language Models.
  https://arxiv.org/html/2312.12705v2.
\item
  Abstract arXiv:2102.07988v2 {[}cs.LG{]} 28 Sep 2021.
  https://arxiv.org/pdf/2102.07988.pdf.
\end{enumerate}

%% file: tables/gpt4t-case.tex
\hypertarget{introduction}{%
\subsection{Introduction}}

The training of large-scale language models, such as GPT-3, BERT, and
T5, requires significant computational resources due to the vast number
of parameters and the size of the datasets involved. To address this
challenge, parallelism techniques have been developed to distribute the
computational workload across multiple processors or machines. The three
main types of parallelism used in training large-scale language models
are data parallelism, model parallelism, and pipeline parallelism. This
report provides an in-depth look at these technologies and their
representative works and frameworks.

\hypertarget{data-parallelism}{%
\subsection{Data Parallelism}}

\hypertarget{definition-and-overview}{%
\subsubsection{Definition and Overview}}

Data parallelism involves splitting the training data across multiple
processors or devices, with each processor holding a complete copy of
the model. Each processor computes the gradients based on its subset of
the data, and the gradients are then aggregated to update the model
parameters.

\hypertarget{representative-works-and-frameworks}{%
\subsubsection{Representative Works and
Frameworks}}

\begin{itemize}
\tightlist
\item
  \textbf{Horovod}: Developed by Uber, Horovod is a distributed training
  framework that uses data parallelism. It is designed to work with
  TensorFlow, Keras, PyTorch, and Apache MXNet. Horovod simplifies the
  process of scaling out deep learning models by providing efficient
  gradient aggregation.
\item
  \textbf{PyTorch Distributed Data Parallel (DDP)}: PyTorch's DDP module
  allows for easy data parallelism implementation. It wraps the model
  during training and manages gradient averaging and synchronization
  across multiple GPUs.
\item
  \textbf{TensorFlow's \texttt{tf.distribute.Strategy}}: TensorFlow
  offers several strategies for distributed training, including
  \texttt{MirroredStrategy} for data parallelism. It synchronizes
  training across multiple GPUs by replicating the model on each device
  and managing gradient updates.
\end{itemize}

\hypertarget{model-parallelism}{%
\subsection{Model Parallelism}}

\hypertarget{definition-and-overview-1}{%
\subsubsection{Definition and
Overview}}

Model parallelism involves splitting the model itself across multiple
processors or devices. Each processor is responsible for a portion of
the model's layers or parameters. This technique is particularly useful
for models that are too large to fit into the memory of a single device.

\hypertarget{representative-works-and-frameworks-1}{%
\subsubsection{Representative Works and
Frameworks}}

\begin{itemize}
\tightlist
\item
  \textbf{GPipe}: Developed by Google, GPipe is a library for
  efficiently training large-scale neural networks using model
  parallelism. It partitions the model into different segments and
  pipelines the execution across multiple accelerators.
\item
  \textbf{Megatron-LM}: Created by NVIDIA, Megatron-LM is a framework
  designed to train transformer-based models using model parallelism. It
  splits the model's layers across multiple GPUs, allowing for the
  training of models with billions of parameters.
\item
  \textbf{Mesh-TensorFlow}: This is a language for distributed deep
  learning, developed by Google, that allows for easy implementation of
  model parallelism. It provides a simple API to distribute the model's
  computation across a mesh of processors.
\end{itemize}

%% file: tables/gpt4t-case2.tex
\hypertarget{pipeline-parallelism}{%
\subsection{Pipeline Parallelism}}

\hypertarget{definition-and-overview}{%
\subsubsection{Definition and Overview}}

Pipeline parallelism combines elements of both data and model
parallelism. The model is split across multiple processors, and the
training data is also split into micro-batches. Each processor works on
a different micro-batch and passes its output to the next processor in
the pipeline.

\hypertarget{representative-works-and-frameworks}{%
\subsubsection{Representative Works and
Frameworks}}

\begin{itemize}
\tightlist
\item
  \textbf{GPipe}: While GPipe is primarily associated with model
  parallelism, it also incorporates aspects of pipeline parallelism by
  pipelining the execution of micro-batches across different model
  partitions.
\item
  \textbf{DeepSpeed}: Developed by Microsoft, DeepSpeed is a deep
  learning optimization library that provides a pipeline parallelism
  implementation called ZeRO-DP. It is designed to optimize memory usage
  and improve scalability when training large models.
\item
  \textbf{FairScale}: An extension of PyTorch, FairScale includes
  several features for distributed training, including pipeline
  parallelism. It allows for the efficient training of large models by
  partitioning the workload across multiple GPUs.
\end{itemize}

\hypertarget{conclusion}{%
\subsection{Conclusion}}

Data parallelism, model parallelism, and pipeline parallelism are
essential techniques for training large-scale language models. Each
technique addresses different challenges associated with the size of the
model and the dataset. Frameworks like Horovod, PyTorch DDP,
TensorFlow's \texttt{tf.distribute.Strategy}, GPipe, Megatron-LM,
Mesh-TensorFlow, DeepSpeed, and FairScale provide the necessary tools
for implementing these parallelism strategies effectively.

As the demand for larger and more complex models continues to grow,
these parallelism techniques and frameworks will play a crucial role in
enabling the development of next-generation AI systems. Researchers and
practitioners must carefully choose the appropriate parallelism strategy
based on the specific requirements of their models and the available
computational resources. With ongoing advancements in distributed
computing and parallel processing, we can expect to see further
improvements in the efficiency and scalability of large-scale language
model training.